\documentclass[12pt,a4paper]{article}

\usepackage[pdftex]{graphicx}
\usepackage{amsmath}
\usepackage{amsopn,amscd,amsthm,amssymb,xypic,rotating}
\usepackage{thmtools}
\usepackage[utf8]{inputenc}
\usepackage{mathtools}
\usepackage{hyperref}
\usepackage{tablefootnote}

\newcommand{\cb}          {\begin{tabbing}MMMMM\=MM\=MM\=MM\=MM\=MM\=MM\=MM\=MM\=MM\= \kill}
\newcommand{\ce}          {\end{tabbing}}

\newcounter{remark}
\newcounter{example}
\usepackage{listings}
\usepackage{color}
\usepackage{soul}
\usepackage{etoolbox}

\definecolor{dkgreen}{rgb}{0,0.6,0}
\definecolor{gray}{rgb}{0.5,0.5,0.5}
\definecolor{mauve}{rgb}{0.58,0,0.82}
\definecolor{lightgray}{rgb}{0.7,0.7,0.7}

\newcommand{\itemtitle}[1] {{\bf{#1}} - }

\newtoggle{annotations}
\toggletrue{annotations}

\title{BiRating - Iterative averaging on a bipartite graph of Beat Saber scores, player skills, and map difficulties}
\author{Juan Casanova}

\begin{document}

\maketitle

Difficulty estimation of Beat Saber maps is an interesting data analysis problem and valuable to the Beat Saber competitive scene. We present a simple algorithm that iteratively averages player skill and map difficulty estimations in a bipartite graph of players and maps, connected by scores, using scores only as input. This approach simultaneously estimates player skills and map difficulties, exploiting each of them to improve the estimation of the other, exploitng the relation of multiple scores by different players on the same map, or on different maps by the same player. While we have been unable to prove or characterize theoretical convergence, the implementation exhibits convergent behaviour to low estimation error in all instances, producing accurate results. An informal qualitative evaluation involving experienced Beat Saber community members was carried out, comparing the difficulty estimations output by our algorithm with their personal perspectives on the difficulties of different maps. There was a significant alignment with player perceived perceptions of difficulty and with other existing methods for estimating difficulty. Our approach showed significant improvement over existing methods in certain known problematic maps that are not typically accurately estimated, but also produces problematic estimations for certain families of maps where the assumptions on the meaning of scores were inadequate (e.g. not enough scores, or scores over optimized by players). The algorithm has important limitations, related to data quality and meaningfulness, assumptions on the domain problem, and theoretical convergence of the algorithm. Future work would significantly benefit from a better understanding of adequate ways to quantify map difficulty in Beat Saber, including multidimensionality of skill and difficulty, and the systematic biases present in score data.

\tableofcontents

\section{Introduction}

Beat Saber\footnote{\url{https://beatsaber.com/}} is a virtual reality rhythm game developed by Beat Games and released in 2018. Players play {\emph{maps}} of {\emph{songs}} that present the player with notes timed to and representing the music that they must cut in specific directions to score, among other gameplay mechanics. Maps for faster songs often have faster patterns that require the player to react more quickly, and other maps have more complex patterns that are more difficult to understand and play. Beat Saber has rich scoring mechanics\footnote{\url{https://bsmg.wiki/ranking-guide.html}}, and a healthy competitive scene with upwards of 60,000 active ranked players in the second half of 2024\footnote{Verified using BeatLeader's search functionality at \url{https://beatleader.com/ranking/1?mapsType=all&recentScoreTime=1721433600}}, which can be observed through its two most important ranked leaderboards: ScoreSaber\footnote{\url{https://scoresaber.com/}} and BeatLeader\footnote{\url{https://beatleader.xyz/}}. Competitive Beat Saber focuses on large pools of {\emph{custom}} maps made by members of the community ({\emph{mappers}}) that undergo a ranking process\footnote{\url{https://beatleader.wiki/en/ranking/Ranking-your-map}} to ensure their competitive viability. For example, as of January 2025, BeatLeader offers over 3500 different ranked maps\footnote{\url{https://beatleader.xyz/leaderboards}}.

Because of the diversity of maps and map difficulty, a key question that ranking leaderboards need to address is: {\emph{How do you compare scores on different maps to determine which one indicates a higher player skill?}} This question is normally reduced to the question of quantifying the difficulty of a map. In general, there is an understanding in the ranked community that Beat Saber skill is not unidimensional, with different players possessing different playstyles and skillsets that allow them to perform better at some maps and worse at others. For example, {\emph{speed}} players will perform well on fast but simple ({\emph{speed}}) maps, where {\emph{tech}} players will perform well on complex but slower ({\emph{tech}}) maps. However, due to the large dimensionality of the variety in maps and playstyles, the problem of accurately quantifying difficulty of maps is complex and not completely understood. This understanding matches existing literature on the subject of difficulty estimation in videogames \cite{measuring_difficulty_single_player_games,measuring_difficulty_platform_games}.\\

ScoreSaber and BeatLeader each use their own machine learning / hybrid algorithms to calculate the difficulty ratings of maps. For example, BeatLeader calculates three different ratings on each map, based on the placement of notes and other objects in the map, each of which contributes in a different way to PP (a measure of how good a certain score on a certain map is):

\begin{itemize}

\item \itemtitle{Pass rating} Indicates how difficult the map is to pass (play without failing). This gives a flat PP to any player that passes the map.
\item \itemtitle{Tech rating} Indicates how complex the map is, especially in terms of the difficulty in understanding it ({\emph{reading it}}). This gives a small amount of PP based on the score the player obtained in the map.
\item \itemtitle{Acc rating} Indicates how difficult it is to achieve a high score (acc) on each of the individual notes of the map, based on their positioning, angle, and context. This gives the majority of the PP obtained by the player, depending on the score the player obtained in the map.

\end{itemize}

Pass and Tech rating are calculated based on rules\footnote{\url{https://github.com/LackWiz/ppCurve}}, while Acc rating uses custom neural networks trained on replays of good players on ranked maps to predict the score that players will likely obtain on each of the individual notes in the map based on their context and parameters\footnote{\url{https://github.com/BeatLeader/beatsaber-replays-ai-2} or \url{https://github.com/DziugasRam/bs-replays-ai-api}}.\\

Iterative algorithms on graph data structures to extract underlying knowledge are widely used \cite{survey_distributed_graph_pattern_matching, survey_distributed_graph_algorithms}. This paper describes a novel iterative averaging algorithm on the graph of maps, players, and scores, to estimate the difficulty of maps and the player skill of players simultaneously based entirely on the relative scores of different players on different maps. More specifically, we can define a bipartite graph with maps and players as the two classes of nodes, and edges representing the best score of the player on that map. This graph allows us to consider the direct relation between multiple scores by different players on the same map, between multiple scores by the same player on different maps, and the transitive relations that can be inferred from this. Conceptually, this graph contains enough information to determine both which maps are more difficult, and which players are better. We need only to extract it. We designed, implemented, and evaluated an iterative averaging algorithm that does this.\\

The remainder of this document is structured as follows.

\begin{itemize}

\item Section \ref{algorithm_description} describes the mathematical principles of the algorithm and some of its properties.
\item Section \ref{data_preparation_implementation} explains the data preparation conducted for this work and relevant details about the implementation of the algorithm.
\item Section \ref{results} presents, discusses, and evaluates the results of the implementation of the algorithm in terms of its ability to produce a good estimation of map difficulty.
\item Section \ref{limitations} discusses the most important known limitations of this approach.
\item Section \ref{related_future_work} discusses some related and future work and its relation to the work presented in this document.
\item Finally, section \ref{conclusions} offers some general conclusions.

\end{itemize}

\section{Algorithm description}
\label{algorithm_description}

This section describes the core conceptual and mathematical principles behind the algorithm. The actual implementation details, along with additional choices and manipulations of the data that were carried out in practice, and their motivation, are explained in \S \ref{data_preparation_implementation}. Here, we stick to the main idealized aspects.\\

\subsection{Bilinear relationship between player skills, map ease, and scores}

The core principle of the algorithm is to define a relationship between {\emph{player skill}} ($p$), {\emph{map difficulty}}, and {\emph{scores}} ($s$). This allows us to calculate the estimated value of any of these three values if we are given the other two. In order to simplify the maths, we replace map difficulty with map {\emph{ease}} ($e$), which can be seen as the inverse of difficulty: a map with higher ease is easier to score well in. We use a simple bilinear relationship between these to enable iterative averaging to work well.\\

\begin{equation}
\label{bilinear_relation}
s = p \cdot e
\end{equation}

For example, if a player has player skill $p = 1.5$ and a map has ease $e = 0.5$, we expect that player to set a score with value $s = 1.5 \cdot 0.5 = 0.75$ on that map. We can also work the formula backwards, for example, if a map has ease $e = 0.25$ and a player sets a score with value $s = 1$ on it, this gives us an estimation of that player's skill as $p = s / e = 1 / 0.25 = 4$.\\

\subsection{Ensuring linearity of values}
\label{ensuring_linearity}

The assumption of bilinearity is mainly for simplicity purposes, and is important for the success of the algorithm, though we expect that other kinds of relationships could also be made to work in similar algorithms. More important is the right choice of the representation of player skill, map ease, and score values. Our data consists of map scores in Beat Saber. While there are multiple ways to represent this, the typical way in which the Beat Saber community looks at them is also useful for our purposes: the score is a percentage representing what proportion of the maximum possible score on a map the player obtained. We will represent this as a value between 0 and 1. 

However, the distribution, and more importantly, the perception of difficulty of scores is far from being uniformly distributed across the $[0,1]$ interval. Here are some general ranges of expected scores:

\begin{itemize}

\item Scores below 0.5 almost universally denote a fail on the map and are extremely poor.
\item A score of 0.8 is still considered relatively poor and most players can achieve this on maps that are approachable to them.
\item A score of 0.9 indicates a decent play depending on the difficulty of the map but for most maps, many players will be able to achieve this.
\item A score of 0.96 is often seen as the gold standard of a good play, with anything above it being a significantly good play. While this definitely depends on the map, 0.96 is frequently seen as the inflection point where increasing score becomes significantly more difficult.
\item A score of 0.98 is very difficult to achieve even on the easiest maps and only the best players are able to reach this even on the simplest maps.
\item A score of 0.99 is extremely high and only the best players on the easiest maps with a lot of practice and iteration can achieve such scores.
\item There is only a single score of 1 (100\%) in the entirety of the ranked leaderboards in BeatLeader, by a high level player on what is considered to be the easiest ranked map.

\end{itemize}

\begin{figure}
\caption{\label{PP_curve}BeatLeader's PP curve on a ranked map.}
\includegraphics[width=0.5\textwidth]{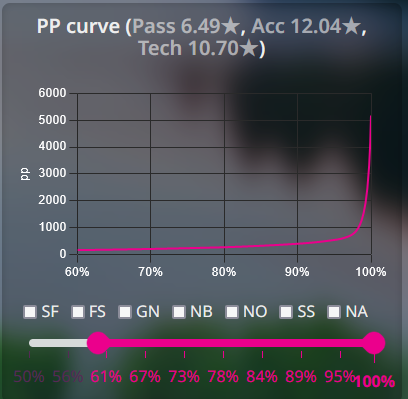}
\centering
\end{figure}

This perception can be visualized in the way in which the current BeatLeader algorithm rewards scores on maps, also known as ``the PP curve''. See figure \ref{PP_curve}.

In order to make our bilinearity assumption work well, we need a way to value scores, skills, and map ease that is linear. Therefore, we need to transform the percentage scores into a different representation that behaves more linearly. Since the score on a map is between 0 and 1 and the value of a play is monotonically increasing, it makes sense to look at it as a probabilistic distribution and use the inverse of the cumulative distribution to estimate the value of the score. Moreover you can interpret a score on a map as the probability of making fewer than a certain number of mistakes in the map. We discuss this approach further and the explicit formulas that we considered and implemented to transform scores into score values in \S \ref{transformation_scores}. For this section, we merely state that the score value $s$ will be calculated in such a way from the scores on maps that it is as close to linear as possible. That is, that a score value that is double of another will represent a twice as valuable score.

We do not need to make any such assumption for player skills or map ease, as we do not use any reference scale for these. We use the score values to estimate player skill and map ease values, and therefore the scaling of the scores will be the one to determine the scaling of map ease and player skill.\\

\subsection{Iterative averaging}
\label{iterative_averaging}

Another core principle of the algorithm is {\emph{iterative averaging}}. If we have an estimation of map ease for each map and know the value of the scores that a certain player has set, we can estimate that player's skill as the average of the estimated skill that each of those scores represents.

\begin{equation}
\label{player_skill_averaging}
\bar{p}  = {{1} \over {n}} \sum\limits_{i = 1}^{n} {{s_i} \over {e_i}}
\end{equation}

Conversely, if we have an estimation of player skill for each player and know the value of the scores on a certain map, we can estimate that map's ease as the average of the estimated ease that each of those scores represents.

\begin{equation}
\label{map_ease_averaging}
\bar{e}  = {{1} \over {n}} \sum\limits_{i = 1}^{n} {{s_i} \over {p_i}}
\end{equation}

The algorithm works by alternating averaging steps of each of these two kinds, updating player skils and map ease based on the scores and their relations to each other.\\

\subsection{Error}
\label{error_section}

One of the desired properties of this algorithm is that the accuracy of the player skills and map ease will increase with each iteration. In order to discuss this at all, we need to usefully define the accuracy of a certain set of values. Given a set of estimations of player skill and map ease values, we can consider what scores those would predict and compare them with the actual scores we observed. We call this the {\emph{error}} of each of the scores, and then define the error of the model to be the mean absolute error of the error of each score $s$ that a player with skill $p$ set on a map with ease $e$.\\

\begin{equation}
\label{error}
\epsilon(s) = |s -  p \cdot e|
\end{equation}

And the average error of the model:

\begin{equation}
\label{average_error}
\bar{\epsilon} ={{1} \over {n}} \sum\limits_{i=1}^{n} \epsilon(s_i)
\end{equation}

There are a few important implicit choices here. First, we note that we use the mean absolute error (MAE) rather than the root mean square error (RMSE) or other similar measures. This is because the error of our model is exclusively an evaluation metric and not an explicit goal of the algorithm, and therefore we do not care about its smoothness properties, which is one of the main reasons to favour RMSE over MAE. However, using the RMSE could make sense as well, for example, to give more weight to outliers with large errors. Ultimately this choice can be exchanged with only moderate consequences in, for example, the optimization of model hyperparameters, and is not a core element affecting the algorithm itself. Second, we do not measure the error in proportion to the value of the score, or mean relative error (MRE). Arguably this could make scores with larger values have more relevance to the error. However, there are two reasons why we favour the MAE: it preserves the linear properties that our general model for relationship between skills, ease, and scores has; and due to the domain problem, it makes sense to give higher value scores have a larger role in evaluating and optimizing the algorithm than small value scores. Competitiveness focuses primarily on the top players achieving the top scores, and we care more about getting those right. However, similar to RMSE, MRE could also be a reasonable metric to evaluate the algorithm under a different scope.\\

A good measure of the correct working of the algorithm would be that the MAE is reduced on each iteration. We discuss the theoretical aspects around this in \S \ref{convergence} and the results in our implementation in \S \ref{results}. We note that the error does not affect the workings of the algorithm, it is just a tool to measure the results.

\subsection{Convergence}
\label{convergence}

Ideally, we would want theoretical proof that the iterative averaging of player skill and map ease values is convergent to a minimum error state. Plenty of results on similar iterative averaging problems follow similar structures \cite{convergence_speed_distributed_consensus,averaging_improve_convergence,average_convergence_iterative_projection,iterative_averaging_saddle_point_problems}. However, we have been unable to find general results that we were  able to apply to our problem. Moreover, our experimental results explained in \S \ref{results} seem to indicate that the method does not exactly converge to a minimum error state. However, there are still some properties of the process, both theoretical and empirically observed, that are relevant to discuss, as well as highlight the theoretical conditions that we are aware would affect this result.\\

First, we note that any state of the model in which all score predictions are perfect ($\bar{\epsilon} = 0$) is a fixed point. This is trivial because if all score predictions are perfect, then the predicted score for each map will be exactly the observed scores, and because we are using the same formula to update estimated player skill and map ease values, for each of these scores, the predicted player skill value will be exactly the current player skill value, and similarly for map ease values. Therefore, the update will not change any values and will remain fixed.

However, in practice it is highly unlikely that an observed dataset will have such a possible state to begin with. It is enough that two players achieved reversely ordered scores on two different maps to ensure that it is not possible to find a fixed point of the system with $\bar{\epsilon} = 0$.\\

We can consider the minimum possible error that a set of observed scores can have within all possible estimations of player skill and map ease values\footnote{Note that the space of possible player skill and map ease values is compact and therefore it will contain a minimum.}. Write ${\bar{\epsilon}}_{\min}$. It would be reasonable to expect that a state with minimum error would be a fixed point. The intuitive argument is that the iteration is doing a best attempt at finding the best way to explain the scores by using averages, and therefore it does not seem expected to have the process increase the error. However, we have not found a theoretical proof of this. Moreover, as described in \S \ref{results}, the empirical process seems to consistently enter a post-optimization regime in which the error slowly increases\footnote{Although there may be other implementation-specific explanations for this phenomenon.}. We cannot at the present  time offer a satisfactory and complete explanation of the reasons behind this, but we conjecture that when certain sets of scores are relevantly incompatible, the iterative averaging may enter an unstable oscillatory regime in which each iteration overcompensates the error on some scores by increasing the error on others. Nonetheless, the process still empirically behaves as convergent, but does not converge to a minimum error state, though it does to a low error state.\\

More in general, the most likely approach to prove convergence of our process would be the Banach fixed-point theorem \cite{wiki_banach_fixed_point}. In order for the theorem to be applicable, we would need to show:
\begin{itemize}
\item The existence of fixed points.
\item Contractiveness of the iterative process.
\end{itemize}

We have not been able to prove the existence of fixed points for every set of possible scores, or characterize the conditions on the set of scores that would guarantee the existence of fixed points, nor the contractiveness of the iterative process. Conceptually, it would be reasonable to expect that a method based on averaging would be contractive, as it explicitly moves the system towards more regularized states, though it is possible that it has an oscillatory component. Empirical results are more promising, though, and are discussed in \S \ref{results}.\\

To be precise, we have empirical evidence (but no proof) for the following {\bf{conjectures}}:

\begin{itemize}
\item The process is always convergent.
\item The process converges to a low error state.
\item The differences between the fixed points and the minimum error states are caused by inherent incompatibilities between scores  that cause small scale oscillations.
\end{itemize}

\subsection{Summary}

\begin{itemize}

\item We define a bilinear relationship between player skills, map ease, and scores defined by equation \ref{bilinear_relation}.

\begin{equation*}
s = p \cdot e
\end{equation*}

\item We ensure linear behaviour of the values by re-scaling scores using probability distribution interpretations of them.

\item We apply iterative mutual averaging to estimate player skill and map ease values, based on each other and using the observed scores, following equations \ref{player_skill_averaging} and \ref{map_ease_averaging}.

\begin{equation*}
\bar{p}  = {{1} \over {n}} \sum\limits_{i = 1}^{n} {{s_i} \over {e_i}}
\end{equation*}

\begin{equation*}
\bar{e}  = {{1} \over {n}} \sum\limits_{i = 1}^{n} {{s_i} \over {p_i}}
\end{equation*}

\item We use mean absolute error to evaluate the results of the processs and observe its convergence properties. These are defined by equations \ref{error} and \ref{average_error}.

\begin{equation*}
\epsilon(s) = |s -  p \cdot e|
\end{equation*}

\begin{equation*}
\bar{\epsilon} = {{1} \over {n}} \sum\limits_{i=1}^{n} \epsilon(s_i)
\end{equation*}

\item While we have not been able to prove convergence, we have investigated the conditions and phenomena that may be affecting the convergence conditions. The process empirically exhibits convergent behaviour towards low error values, though they are not minimum error values.

\end{itemize}

\section{Data preparation and implementation}
\label{data_preparation_implementation}

The implementation used for this research is available on GitHub\footnote{\url{https://github.com/Undeceiver/BiRating}}. The data used consists on all scores for all ranked maps on BeatLeader\footnote{\url{https://beatleader.xyz/}} up to 02/11/2024. This was obtained directly from BeatLeader's administrator, but it can also be extracted using BeatLeader's public API\footnote{\url{https://api.beatleader.xyz/swagger/index.html}}. This contains over 2.5 million scores for over 3500 maps. We note that only the best score for each player and map is stored by BeatLeader. The same player may not have two scores on the same map.\\

\subsection{Data preparation}
\label{data_preparation}

Preliminary data preparation included a number of data cleaning and filtering steps:
\begin{itemize}
\item Scores below 0.75 were removed. These scores often misrepresent the difficulty of maps and frequently correspond to players who did not even try to play the map, were playing badly on purpose, or had other similar issues. They are outliers that reduce the accuracy of the method.
\item Scores (only 1) of exactly 1 were removed. For some of our mathematic manipulations, a score of 1 generates exceedingly high values that produce unstable iterative behaviour. Since there was only 1 such score, it was removed.
\item Out of all of these, we preserved only the 35\% most recent scores. This was done to improve the quality of the result as player skill improves over time, which our algorithm does not account for. It is likely that older scores do not accurately represent the skill level of the player, and therefore the difficulty of the map, making older maps appear as inherently harder. This left us with just over 850,000 scores. There are over 3 years of scores stored in BeatLeader, so we can expect this to be approximately a year of scores.
\item Scores in BeatLeader can include {\bf{modifiers}}. These are modifications to the gameplay that can make maps easier or harder, modifying the score. We adjusted the score values using the following table of modifiers (only modifiers that actually change the score were included). Note that this table is designed to be an underestimation of the difficulty of maps under the modifiers, giving unmodified scores more value. This is a known challenge to the accuracy of our problem, but unfortunately removing modified scores altogether reduced the amount of scores for some players too much. Section \S \ref{limitations} discusses this in more depth.\\
\begin{tabular}{ |c|c|c| }
\hline
{\bf{Abbreviation}}&{\bf{Modifier}}&Score multiplier\\
\hline
SF&Super Fast Song&1.05\\
\hline
SA&Strict Angles&1.02\\
\hline
NO&No Obstacles&0.5\\
\hline
SS&Slow Song&0.65\\
\hline
FS&Fast Song&1.02\\
\hline
NB&No bombs&0.5\\
\hline
NA&No Arrows&0.5\\
\hline
OP&Out of Platform&0.5\\
\hline
\end{tabular}
\end{itemize}

\subsection{Mathematical transformations of score values}
\label{transformation_scores}

As discussed in \S \ref{ensuring_linearity}, the algorithm relies on linearity of the value of scores to work properly due to iterative averaging, but the scores are not inherently linear and biased around the 0.8-0.95 area. In order to transform this into a linear representation of scores, we used a probabilistic interpretation of the score. While there is some reasoning behind this transformation, the main guiding element here is a good outcome where the resulting scores seem to behave linearly in terms of score difficulty / skill required. Note that this transformation transforms scores independently of map or player parameters, taking numbers in the $[0,1]$ interval and producing numbers in the $[0,100]$ interval that behave more linearly with respect to difficulty.\\

In order to do this, we leverage two standard probabilistic distributions.

\begin{itemize}

\item The beta distribution family represents distributions in the $[0,1]$ range with a lot of flexibility in the parameters. This is used to capture the fundamental shape of the original curve of scores, by applying the beta distribution (with appropriately chosen parameters) cumulative distribution function to the original score to obtain an approximate percentile value that the score represents, representing scores on the $[0,1]$ range in a mathematically more meaningful way.

\item The higher a score, the harder it is to improve it further. In some sense, high scores relate to the {\emph{perfect execution}} of the map and count how many errors the player has made when playing (missing notes, not hitting notes accurately, etc.). This makes an exponential distribution a reasonable way to evaluate how much value higher scores have as being increasingly harder to improve the higher the score is. One important issue with an exponential distribution is that it is unbounded. We can solve this by using a truncated exponential distribution. We apply the percentile function to the percentile value obtained above to calculate a final score value using a truncated exponential distribution with predefined parameters.

\end{itemize}

The results of this transformation were inspected and analysed to observe a more linear behaviour with a good distribution of score values across the entire range, that more accurately reflects the value of scores in a linear way. Moreover, we included the parameters of the beta and truncated exponential distributions described above as part of a {\bf{hyperparameter grid search}} (see \S \ref{hyperparameter_grid_search}) looking to minimize the average error in the predicted scores.\\

\subsection{Selecting scores and error values for improved aggregation}
\label{selecting_aggregation}

A significant challenge to the accuracy of the iterative process is the presence of outliers and unrepresentative scores. While we have tried to deal with this in the data preparation (see \S \ref{data_preparation}), we found it valuable to include data curation in the iterative process itself.

In particular, when calculating the average player skill from map ease values, or map ease values from player skill (see \S \ref{iterative_averaging}), we restrict this to a proportion of the highest scores for each particular player and/or map ease values. This represents the notion, normally acknowledged in rating algorithms, that top scores represent player skill and map difficulty much more accurately than poor scores. A player is best represented by their best scores, and a map is best represented by the scores that the best players set on it. In a competitive environment, this systematic choice improves results by discounting accidents and unrepresentative scores. This also relates to the fact that the original dataset only had the best scores for each player on each map to begin with.

Similarly, when calculating the predictive error of the model (see \S \ref{error_section}), we limit the average error to the lowest half of errors in scores. Once again, this excludes outlier scores that are unrepresentative of the players or maps, preventing them from overblowing the resulting error.\\

Admittedly, these choices can also affect the outcome of the algorithm and of the evaluation, making it look better than it is. However, for the purposes of obtaining good final approximations of map difficulties, this proved to be preferable, and we discuss the qualitative evaluation of the results as well in \S \ref{results}. The particular proportions used were subject to {\bf{hyperparameter grid search}} (see \S \ref{hyperparameter_grid_search}).

\subsection{Hyperparameter grid search}
\label{hyperparameter_grid_search}

In order to try a variety of algorithm hyperparameters to try to improve the results, we conducted several rounds of hyparameter grid search, in each case trying to learn from the results, adjusting minor elements of the algorithm, and deciding other hyperparameter variations to try, eventually arriving at a better combination of hyperparameters.\\

The following are all the hyperparemeters of the algorithm considered:

\begin{itemize}

\item \lstinline|aggregation_topscores_p| - Proportion of best scores that are considered in averaging the player skill  / map ease calculations. \S \ref{selecting_aggregation}.
\item \lstinline|aggregation_topscores_sd_range| - Introduced in later versions. Removes scores with value that is over this number times the standard deviation of the top scores. \S \ref{selecting_aggregation}
\item \lstinline|beta_alpha| - Alpha parameter of the beta distribution to use for the transformation of score values. \S \ref{transformation_scores}.
\item \lstinline|beta_beta| - Beta parameter of the beta distribution to use for the transformation of score values. \S \ref{transformation_scores}.
\item \lstinline|default_rating| - Default (standard) rating for map ease / player skill used to calculate the mean of the used distributions as well as other calculations.
\item \lstinline|error_change_prop| - When the mean average error of the estimated scores changes by less than this proportion in an iteration of the algorithm, the algorithm halts.
\item \lstinline|finish_early| - True or False. When set to True, if the error increases in an iteration of the algorithm, the algorithm reverts to the previous values and halts.
\item \lstinline|truncexp_base_mean| - Mean of the truncated exponential distribution used for the transformation of score values. \S \ref{transformation_scores}. This parameter should not affect accuracy as it only changes the scale of how score values are represented.
\item \lstinline|truncexp_max| - Maximum value of the truncated exponential distribution used for the transformation of score values. \S \ref{transformation_scores}. This parameter should not affect accuracy as it only changes the scale of how score values are represented.

\end{itemize}

We discuss which were varied and used in each of the hyperparameter runs in \S \ref{results}.

\subsection{Summary}

\begin{itemize}

\item We used an initial dataset of over 2.5 million scores on over 3500 ranked maps from BeatLeader, comprising all scores registered on ranked maps up to 02/11/2024.
\item Data was prepared and cleaned by removing uncommonly low scores, perfect scores that would hurt the algorithm's performance, old scores, and adjusting for gameplay modifiers.
\item Score values were adjusted to behave more linearly by using a probabilistic interpretation that reverts the curve shape of the score distribution using a beta distribution, and readjusts it to a value curve using a truncated exponential distribution.
\item During the algorithm, scores used in averaging and errors measured are limited to a proportion of all scores and errors, for the purpose of further limiting the effect of outliers and unrepresentative scores.
\item Hyperparameter grid search is used to improve the performance of the algorithm.

\end{itemize}

\section{Results}
\label{results}

\subsection{Quantitative evaluation}

In this section we discuss the particular values of the hyperparameters tried during hyperparameter grid search and the results, and some additional details about the numerical results of the best hyperparameter combination found. All runs involve 5 crossvalidation runs with a 80/20 train/test split, averaging the error\footnote{Adjusted as described in \S \ref{selecting_aggregation}} across the 5 runs separately on training and test data. Best hyperparameter combination in each run is highlighted in bold.

\subsubsection{Hyperparameter search 1}

\noindent {\bf{Fixed values}}

\begin{itemize}

\item \lstinline|beta_alpha|: $10$
\item \lstinline|beta_beta|: $1.25$
\item \lstinline|default_rating|: $10$
\item \lstinline|finish_early|: False
\item \lstinline|truncexp_base_mean|: $10$
\item \lstinline|truncexp_max|: $100$

\end{itemize}

\noindent {\bf{Grid values}}

\begin{itemize}

\item \lstinline|aggregation_topscores_p| (P): $0.25, 0.5, 0.75, 0.9$
\item \lstinline|error_change_prop| (E): $0.005, 0.001, 0.0002$

\end{itemize}

\noindent {\bf{Results}}

\begin{center}
\makebox[0.1cm]
{
\begin{tabular}{ |c|c|c|c| }
\hline
P&E&{\bf{Training MAE}}&{\bf{Test MAE}}\\
\hline
0.25&0.005&1.212&2.061\\
\hline
0.25&0.001&1.209&1.954\\
\hline
0.25&0.0002&1.207&1.963\\
\hline
0.5&0.005&0.819&1.411\\
\hline
0.5&0.001&0.819&1.395\\
\hline
0.5&0.0002&0.816&1.337\\
\hline
0.75&0.005&0.693&1.234\\
\hline
0.75&0.001&0.69&1.275\\
\hline
0.75&0.0002&0.69&1.27\\
\hline
{\bf{0.9}}&{\bf{0.005}}&{\bf{0.66}}&{\bf{1.202}}\\
\hline
0.9&0.001&0.656&1.241\\
\hline
0.9&0.0002&0.657&1.233\\
\hline
\end{tabular}
}
\end{center}

\noindent {\bf{Comments}}

This run indicates quite clearly that a proportion of 0.9 (90\% of scores) for averaging each run leads to lower error, as well as suggest that a higher threshold for stopping the iteration is better. The latter relates to a phenomenon observed in almost all runs that the error after each iteration initially goes down quite quickly, but then begins to increase slowly after a certain saturation point is reached.

\subsubsection{Hyperparameter search 2}

\noindent {\bf{Fixed values}}

\begin{itemize}

\item \lstinline|beta_alpha|: $10$
\item \lstinline|beta_beta|: $1.25$
\item \lstinline|default_rating|: $10$
\item \lstinline|finish_early|: True
\item \lstinline|truncexp_base_mean|: $10$
\item \lstinline|truncexp_max|: $100$

\end{itemize}

\noindent {\bf{Grid values}}

\begin{itemize}

\item \lstinline|aggregation_topscores_p| (P): $0.5, 0.9$
\item \lstinline|aggregation_topscores_sd_range| (S): $1, 3$
\item \lstinline|error_change_prop| (E): $0.005, 0.001$

\end{itemize}

\noindent {\bf{Results}}

\begin{center}
\makebox[0.1cm]
{
\begin{tabular}{ |c|c|c|c|c| }
\hline
P&S&E&{\bf{Training MAE}}&{\bf{Test MAE}}\\
\hline
0.5&1&0.005&0.77&1.638\\
\hline
0.5&1&0.001&0.77&1.68\\
\hline
0.5&3&0.005&0.837&1.704\\
\hline
0.5&3&0.001&0.838&1.698\\
\hline
{\bf{0.9}}&{\bf{1}}&{\bf{0.005}}&{\bf{0.636}}&{\bf{1.262}}\\
\hline
0.9&1&0.001&0.637&1.305\\
\hline
0.9&3&0.005&0.678&1.396\\
\hline
0.9&3&0.001&0.679&1.527\\
\hline
\end{tabular}
}
\end{center}

\noindent {\bf{Comments}}

This run further confirms 0.9 and 0.005 as best values for \lstinline|aggregation_topscores_p| and \lstinline|error_change_prop| respectively, while indicating that a more restrictive standard deviation range within the topscore selection pays off by improving error.

\subsubsection{Hyperparameter search 3}

\noindent {\bf{Fixed values}}

\begin{itemize}

\item \lstinline|default_rating|: $10$
\item \lstinline|finish_early|: True
\item \lstinline|error_change_prop|: $0.005$
\item \lstinline|truncexp_base_mean|: $10$
\item \lstinline|truncexp_max|: $100$
\item \lstinline|aggregation_topscores_p|: $0.9$
\item \lstinline|aggregation_topscores_sd_range|: $1$

\end{itemize}

\noindent {\bf{Grid values}}

\begin{itemize}

\item \lstinline|beta_alpha| ($\alpha$): $5, 7.5, 10, 14, 18, 25$
\item \lstinline|beta_beta| ($\beta$): $1.05, 1.1, 1.25, 1.5, 2.25$

\end{itemize}

\noindent {\bf{Results}}

\begin{center}
\makebox[0.1cm]
{
\begin{tabular}{ |c|c|c|c|c| }
\hline
$\alpha$&$\beta$&{\bf{Training MAE}}&{\bf{Test MAE}}\\
\hline
5&1.05&0.68&1.341\\
\hline
5&1.1&0.713&1.416\\
\hline
5&1.25&0.818&1.569\\
\hline
5&1.5&0.992&1.908\\
\hline
5&2.25&1.54&3.04\\
\hline
7.5&1.05&0.595&1.197\\
\hline
7.5&1.1&0.628&1.251\\
\hline
7.5&1.25&0.729&1.446\\
\hline
7.5&1.5&0.899&1.744\\
\hline
7.5&2.25&1.416&2.626\\
\hline
10&1.05&0.512&1.022\\
\hline
10&1.1&0.543&1.084\\
\hline
10&1.25&0.636&1.266\\
\hline
10&1.5&0.8&1.614\\
\hline
10&2.25&1.306&2.472\\
\hline
14&1.05&0.392&0.763\\
\hline
14&1.1&0.417&0.824\\
\hline
14&1.25&0.497&0.978\\
\hline
14&1.5&0.641&1.306\\
\hline
14&2.25&1.116&2.155\\
\hline
18&1.05&0.29&0.582\\
\hline
18&1.1&0.311&0.62\\
\hline
18&1.25&0.378&0.764\\
\hline
18&1.5&0.5&1.007\\
\hline
18&2.25&0.925&1.816\\
\hline
{\bf{25}}&{\bf{1.05}}&{\bf{0.164}}&{\bf{0.381}}\\
\hline
25&1.1&0.178&0.4\\
\hline
25&1.25&0.222&0.499\\
\hline
25&1.5&0.306&0.687\\
\hline
25&2.25&0.634&1.295\\
\hline
\end{tabular}
}
\end{center}

\noindent {\bf{Comments}}

The intention of this run was to optimize the beta distribution parameters, and it became clear that a higher value for alpha and a lower value for beta was desirable. This changes the shape of the beta distribution curve to be more heavily biased near $1$, pushing the importance of score improvement closer to a perfect score.

\subsubsection{Hyperparameter search 4}

\noindent {\bf{Fixed values}}

\begin{itemize}

\item \lstinline|beta_alpha|: $25$
\item \lstinline|default_rating|: $10$
\item \lstinline|finish_early|: True
\item \lstinline|error_change_prop|: $0.005$
\item \lstinline|truncexp_base_mean|: $10$
\item \lstinline|truncexp_max|: $100$
\item \lstinline|aggregation_topscores_p|: $0.9$
\item \lstinline|aggregation_topscores_sd_range|: $1$

\end{itemize}

\noindent {\bf{Grid values}}

\begin{itemize}

\item \lstinline|beta_beta| ($\beta$): $1.01, 1.02, 1.05$

\end{itemize}

\noindent {\bf{Results}}

\begin{center}
\makebox[0.1cm]
{
\begin{tabular}{ |c|c|c|c|c| }
\hline
$\beta$&{\bf{Training MAE}}&{\bf{Test MAE}}\\
\hline
1.01&0.153&0.356\\
\hline
{\bf{1.02}}&{\bf{0.156}}&{\bf{0.35}}\\
\hline
1.05&0.165&0.389\\
\hline
\end{tabular}
}
\end{center}

\noindent {\bf{Comments}}

We wanted to test some additional values for the beta distribution, but it seems clear that the improvement is nearly saturated.\\

\subsubsection{Further analysis with optimized hyperparameters}

As a result of this hyperparemeter search, we choose the final hyperparameters:

\begin{itemize}

\item \lstinline|beta_alpha|: $25$
\item \lstinline|beta_beta|: $1.02$
\item \lstinline|default_rating|: $10$
\item \lstinline|finish_early|: True
\item \lstinline|error_change_prop|: $0.005$
\item \lstinline|truncexp_base_mean|: $10$
\item \lstinline|truncexp_max|: $100$
\item \lstinline|aggregation_topscores_p|: $0.9$
\item \lstinline|aggregation_topscores_sd_range|: $1$

\end{itemize}

This choice is not necessarily ideal but it seems clear that most of the improvement attainable through hyperparameter optimization of relevant hyperparameters is saturated. It is relevant to note that in every run the test error has been noticeably higher than the training error, implying that the process is quite susceptible to the particular scores and struggling to find patterns of player skill and map difficulty that enable reliable extrapolation.\\

\begin{figure}
\caption{\label{error_iteration}Error progression during iteration}
\includegraphics[width=0.95\textwidth]{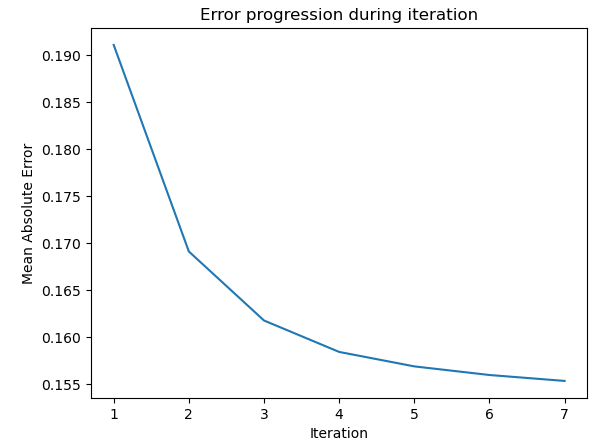}
\centering
\end{figure}

We can see the evolution of the mean absolute error as the algorithm, with the chosen hyperparameters, iterates on the full dataset as prepared. See figure \ref{error_iteration}. The algorithm finishes after 7 iterations due to small variation in the MAE, with an MAE of approximately 0.156. It is clear that the mean absolute error decreases through the iteration and it seems reasonable to assume that the process is convergent at least in this dataset with these hyperparameters, as can be observed by the progression of the error. All other runs of the algorithm that we have carried out exhibit similar behaviour, though in many cases the MAE begins to increase slowly towards the latter iterations, while still behaving in a convergent manner.

\subsection{Qualitative evaluation}

In order to better validate the results of the algorithm in practical terms, we conducted a qualitative validation of results based on presenting them to the BeatLeader ranked community in a way they can understand and compare, and ask them to provide insights and thoughts.\\

In particular, the map ease values resulting from the final run of the algorithm were rescaled using an affine transformation to more directly compare with current BeatLeader map difficulty values. More precisely:

\begin{itemize}

\item The hardest and easiest maps as output by our algorithm correspond to the hardest and easiest map in BeatLeader currently.
\item We adjust map ease values with an affine transformation that guarantees that the difficulty quantification for the hardest and easiest maps is the same in our algorithm results and current BeatLeader's difficulty estimation.
\item All the other maps are therefore linearly interpolated from our map ease value inbetween those.
\item We compare those values with their current values in Beat Leader, and present these in a public website, ordered from hardest to easiest, with a special indication on those maps for which the resulting estimation is significantly different than it currently is in BeatLeader.

\end{itemize}

The resulting page that was shared with BeatLeader ranked players and mappers can be found at \url{https://tinyurl.com/53dm6sbh}.\\

There were approximately 10-20 members of the community that actively engaged with this process, offering thoughts on specific maps, general patterns, and others, over the course of a couple of weeks. The following are some summarized themes that arose from those conversations:

\begin{itemize}

\item Generally, the difficulty estimation ballpark of most maps is consistent between the algorithm and BeatLeader's current algorithm and is reasonable in all cases.
\item Some maps that are renowned in the community for being badly estimated by the current BeatLeader algorithm (underestimating or overestimating their difficulty) seemed to be estimated much better by this algorithm. Examples include \url{https://beatleader.xyz/leaderboard/global/2dd6cxx92/1} (12.15 on BeatLeader, estimated 14.9 by our algorithm), \url{https://beatleader.xyz/leaderboard/global/3963ex92/1} (11.84 on BeatLeader, estimated 10.5 by our algorithm), \url{https://beatleader.xyz/leaderboard/global/23c7bxxx91/1}  (11.08 on BeatLeader, estimated 12.8 by our algorithm).
\item Maps in the middle difficulty range seemed to be estimated worse than the harder and easier ones by our algorithm. This suggests issues with linearity of map difficulty.
\item A certain subset of older maps seemed to be consistently underestimated by our algorithm. We conjecture this is because people have had more time to optimize their scores on those maps, which leads our algorithm to presume the map is easier than it is.
\item Similarly, a certain subset of broadly disliked maps seemed to be consistently overestimated by our algorithm. We conjecture this is because people have not spent as much time optimizing their scores on those maps (because they do not particularly enjoy playing those maps), which leads our algorithm to presume the map is harder than it is.

\end{itemize}

As a summary, the qualitative evaluation validated to a large degree the pragmatic validity of our algorithm, while outlining some important challenges with the approach, particularly around linearity assumptions and the reliance on observed scores, especially due to the lack of context of how said scores were produced. We discuss these topics further in \S \ref{limitations}.

\section{Limitations}
\label{limitations}

In this section we discuss the known limitations of our approach, both of the conceptual algorithm, and of the application to the domain problem of estimating map difficulty in Beat Saber.

\subsection{Data limitations}
\label{data_limitations}

The data used was not captured with the intention of estimating map difficulty or data analysis at all, but rather for the purposes of running the leaderboard. Therefore, there are a number of quality issues with it when used for data analysis.  Only the best score of each player on each map is preserved. Some of the scores were not set while the players were trying their best, but the algorithm assumes they were. Gameplay modifiers (\S \ref{data_preparation}) are difficult to quantify in relation to no modifier scores, which in combination with preserving only the highest score (which can make modified scores override unmodified scores), severely hurts the reliability of the data as a representation of player skill. Further to this, some maps see significantly more attempts and therefore have higher quality scores, which ultimately negatively affects their difficulty estimation by the algorithm.\\

An interesting notion is whether an algorithm like this could be used in a live ranking system to automatically adjust the ranking of players and maps. There are a number of additional issues with this, that come from the potential abuse that it could encourage. For example, bad scores on a map would increase the map's estimated difficulty, which would make good scores on that map a lot more valuable. Players could attempt to intentionally set bad scores on certain maps, either with alternate accounts or for their friends to rank higher. Perhaps even more importantly, the approach inherently requires a minimum volume of scores on each map and for each player to be able to provide a somewhat reliable prediction of their skills and difficulties, which would make new maps or new ranked players extremely unreliable.

\subsection{Domain problem factors}
\label{domain_problem_factors}

As it has been discussed already, different types of players find different types of maps easier or harder. In a sense, player skill and map difficulty are {\emph{multidimensional}}. Our approach inherently assumes that these multiple dimensions can be quantified in a single number. While a certain combination of skills can be encoded in a single number by giving them multiple weights, this adds implicit assumptions to the problem and likely introduces instability to the values when new maps, players, or scores are set.

This also relates to the curve shape problem discussed before. Both the current approach followed by ScoreSaber and BeatLeader, and our approach, assume a uniparameter curve for maps (slightly more complex for BeatLeader with 3 separate sources of difficulty, but ultimately very similar). This means that it is not possible for the approaches to consider that it might be possible, for example, that in map A achieving a 90\% score is very easy, but achieving a 95\% score is very hard, whereas in map B achieving a 90\% score is moderately hard, but it is not that much harder to achieve a 95\% score from there. This very much does happen in Beat Saber, especially when the difficulty of a map relates to three separate things: Passing the map, passing the map with Full Combo (not missing any note), and achieving a high accuracy score on all the notes in the map. These difficulties are inherently independent from each other, and probably can be unrolled into multiple other dimensions that we are, as of now, unaware of. By using uniparameter difficulty and skill estimations, we are blocking the possibility of our algorithm to account for these variations.

Some of this is further discussed in \S \ref{related_future_work}.

\subsection{Inherent algorithm limitations}
\label{inherent_algorithm_limitations}

In \S \ref{convergence} we discussed the mathematical convergence properties of the algorithm. We explained that we could not achieve a proof of convergence or of the existence of fixed points for the algorithm or a characterization of the conditions under which this would be true. However, in \S \ref{results} we showed that all runs of the algorithm that we have carried out on real data has exhibited convergent behaviour. The relation to a minimum inherent error is also not entirely clear, or whether fixed points would necessarily be minimum error points or no. There is a very real possibility that the algorithm can be oscillating in general and/or that certain patterns in the data can systematically hurt the validity of the results. While the results are practically useable and solid, these possibilities put in question the general capabilities of the basic algorithmic approach.

\section{Related and future work}
\label{related_future_work}

It is widely recognized that difficulty estimation in videogames is useful for many reasons \cite{measuring_difficulty_single_player_games,measuring_difficulty_platform_games}. However, to the best of the author's knowledge, most existing approaches to difficulty estimation fall under three main categories:

\begin{itemize}

\item {\emph{Ad hoc solutions}} for specific games that leverage specific knowledge about the game and its relation to difficulty \cite{puzzle_difficulty_estimation, difficulty_sudoku, measuring_difficulty_platform_games}. That is, ad hoc rule-based approaches. The method used for Tech and Pass rating in BeatLeader's currently used algorithm would fall under this category.

\item {\emph{Predictive models}} that aim to predict the performance of a player in a certain level or task as a means to predict difficulty \cite{puzzle_difficulty_simulated_data}. While our approach is most related to these, and can be used to predict the performance of players in maps, there is a difference in the explicit use of the transitive relations between maps and other maps players have played, players and other players that have played the same maps, and further degrees of this. Existing predictive models often focus on either individual players, individual groups of players, or a representative player, instead of analysing the relations between them.

\item {\emph{Agent-based simulations}} that expand predictive models by generating simulated data using agents that attempt to play the game \cite{puzzle_difficulty_simulated_data}. The method used for Acc rating in BeatLeader's falls between predictive models and this approach, as it has been used to train a model that can simulate actual plays on a map by a virtual agent\footnote{\url{https://muffnlabs.de/cyberramen}}.

\end{itemize}

This approach is different to previous attempts to quantify difficulty in Beat Saber maps specifically primarily in that it utilizes scores exclusively to determine difficulty, rather than note and object placement in the map. This means both that the algorithm uses a significantly smaller amount of data to train, and that it may focus exclusively on the observed patterns of difficulty in the scores by different players to quantify and compare more objectively the relative difficulty of maps. However, it also means that the applicability of the algorithm has a number of important limitations, discussed in \S \ref{limitations}. Furthermore, this algorithm is unsupervised, using observations of scores to estimate skills and difficulties, rather than requiring labelled data about skills or difficulties.

This approach is different to other iterative averaging algorithms in bipartite graphs \cite{consensus_bipartite_graphs, iterative_algorithm_priority_pairwise, averaging_process_bipartite_graphs} in the particularities of the mathematical problem. In particular, in this problem there is a tension between player skills and map difficulties, where a better score could be explained {\emph{either}} by an easier map or a better player, and a better player skill increases the perceived difficulty of the map, and viceversa. More importantly, the edges (scores) on the graph do not represent a strength of the relation between players and maps, but rather an observation that needs to be explained and can be explained in multiple ways. In typical recommender systems and other problems for which iterative averaging algorithms are used, the underlying situation is one of similarities and relatedness. For example, \cite{consensus_bipartite_graphs} is concerned with the consensus problem, which involves all nodes in the network agreeing on a single value. Similarly, \cite{iterative_algorithm_priority_pairwise} tries to rank nodes based on priority, in which every adjacent node contributes equally and in which higher priority neighbours increase the priority of the nodee. By contrast, in our problem, all players, including high skill players, will obtain better scores on easier maps, and a better player will obtain better scores on all maps. In other words, higher values of the edges (scores) do not represent a stronger connection between the nodes, merely an observation to be explained. This drastically changes the underlying maths, but can still be approached using an iterative averaging algorithm.\\

While the algorithm and the data used in this paper could be further improved to obtain better results and more thorough conclusions, we believe that new elements or variations on the approach are necessary to achieve significantly better results and overcome some of the discussed limitations (\S \ref{limitations}). One particular direction that the author has considered exploring in the near future is the curve shape issues (of the curve of difficulty/value of a score as compared to \% score on the map) and characterizing map difficulty as multiparameter curves rather than single parameter curves. Multiple approaches, including optimisation and statistical techniques, can be used to optimize multi-parameter curves to a score population on a map, and might produce significantly better results than the ones obtained here. Identifying a suitable parametrical family of curves that successfully work to characterize the difficulty of all or most maps could be a really important stepping stone in improving other approaches to improving difficulty estimation in Beat Saber, for example by giving a better gold standard (based on scores) that predictive models and agent-based simulations can aspire to in new maps with no scores.

\section{Conclusions}
\label{conclusions}

Difficulty estimation in video games is an interesting problem with plenty of applications. In Beat Saber, the competitive scene largely depends on and benefits from accurate methods of difficulty estimation. We presented an interesting but quite simple approach to estimating map difficulty in Beat Saber using player scores but acknowledging the relation between multiple players on the same map and multiple maps played by the same player, transitively. This is slightly different, but closely related, to other approaches in both game difficulty estimation and iterative algorithms in graphs; and presents some interesting particularities and ideas. The implementation is also relatively simple and runs successfully on large amounts of scores on user hardware. The results produced are quite successful, and in some aspects better than existing approaches, but also highlight some limitations in the approach, such as difficulty weighing existing data and making assumptions about it, and being unaware of the context in which it was produced. Some of these difficulties would persist with extensions and improvements to the approach, and therefore highlight the need for additional approaches or significant changes to some of the assumptions of the algorithm. We believe that exploring the role of multiparameter curve families that represent the distribution and value of scores in maps would be a significant step in improving difficulty estimation approaches.

\section{Additional details}

ChatGPT was used for the following purposes while carrying out this work:

\begin{itemize}

\item Support in finding related approaches and references easily and give a basis for a comparison with our approach.
\item Support in understanding the convergence properties of our algorithm and the potential reasons that the algorithm may or may not be convergent.

\end{itemize}

Everything else was produced independently by the author, including the entirety of the text of the paper and all source code.

\bibliographystyle{plain}
\bibliography{bibliography}

\begin{thebibliography}{10}

\bibitem{average_convergence_iterative_projection}
Ya.~I. Alber.
\newblock On average convergence of the iterative projection methods.
\newblock {\em Taiwanese Journal of Mathematics}, 6(3):323--341, 2002.

\bibitem{measuring_difficulty_single_player_games}
Maria-Virginia Aponte, Guillaume Levieux, and Stephane Natkin.
\newblock Measuring the level of difficulty in single player video games.
\newblock {\em Entertainment Computing}, 2(4):205--213, 2011.
\newblock Special Section: International Conference on Entertainment Computing
  and Special Section: Entertainment Interfaces.

\bibitem{survey_distributed_graph_pattern_matching}
Sarra Bouhenni, Sa\"{\i}d Yahiaoui, Nadia Nouali-Taboudjemat, and Hamamache
  Kheddouci.
\newblock A survey on distributed graph pattern matching in massive graphs.
\newblock {\em ACM Comput. Surv.}, 54(2), February 2021.

\bibitem{averaging_process_bipartite_graphs}
Pietro Caputo, Matteo Quattropani, and Federico Sau.
\newblock Cutoff for the averaging process on the hypercube and complete
  bipartite graphs.
\newblock {\em Electronic Journal of Probability}, 28(none), January 2023.

\bibitem{iterative_averaging_saddle_point_problems}
Yuan Gao, Christian Kroer, and Donald Goldfarb.
\newblock Increasing iterate averaging for solving saddle-point problems, 2020.

\bibitem{consensus_bipartite_graphs}
Martin Kenyeres and Jozef Kenyeres.
\newblock Distributed average consensus algorithms in d-regular bipartite
  graphs: Comparative study.
\newblock {\em Future Internet}, 15(5), 2023.

\bibitem{puzzle_difficulty_simulated_data}
Jeppe~Theiss Kristensen and Paolo Burelli.
\newblock Difficulty modelling in mobile puzzle games: An empirical study on
  different methods to combine player analytics and simulated data.
\newblock {\em International Journal of Computer Games Technology},
  2024(1):5592373, 2024.

\bibitem{averaging_improve_convergence}
W.~Robert Mann.
\newblock Averaging to improve convergence of iterative processes.
\newblock In M.~Zuhair Nashed, editor, {\em Functional Analysis Methods in
  Numerical Analysis}, pages 169--179, Berlin, Heidelberg, 1979. Springer
  Berlin Heidelberg.

\bibitem{survey_distributed_graph_algorithms}
Lingkai Meng, Yu~Shao, Long Yuan, Longbin Lai, Peng Cheng, Xue Li, Wenyuan Yu,
  Wenjie Zhang, Xuemin Lin, and Jingren Zhou.
\newblock A survey of distributed graph algorithms on massive graphs.
\newblock {\em ACM Comput. Surv.}, 57(2), October 2024.

\bibitem{measuring_difficulty_platform_games}
Fausto Mourato and Manuel Pr{\'o}spero~dos Santos.
\newblock Measuring difficulty in platform videogames.
\newblock In {\em 4. {\textordfeminine} Confer{\^e}ncia Nacional
  Interac{\c{c}}{\~a}o humano-computador}, 2010.

\bibitem{convergence_speed_distributed_consensus}
Alex Olshevsky and John~N. Tsitsiklis.
\newblock Convergence speed in distributed consensus and control, 2009.

\bibitem{puzzle_difficulty_estimation}
Satoshi Ono, Ryuji Miyamoto, Shigeru Nakayama, and Kazunori Mizuno.
\newblock Difficulty estimation of number place puzzle and its problem
  generation support.
\newblock In {\em 2009 ICCAS-SICE}, pages 4542--4547, 2009.

\bibitem{difficulty_sudoku}
Radek Pelánek.
\newblock Difficulty rating of sudoku puzzles: An overview and evaluation,
  2014.

\bibitem{iterative_algorithm_priority_pairwise}
Haomin Wang, Gang Kou, and Yi~Peng.
\newblock An iterative algorithm to derive priority from large-scale sparse
  pairwise comparison matrix.
\newblock {\em IEEE Transactions on Systems, Man, and Cybernetics: Systems},
  52(5):3038--3051, 2022.

\bibitem{wiki_banach_fixed_point}
{Wikipedia contributors}.
\newblock Banach fixed-point theorem --- {Wikipedia}{,} the free encyclopedia,
  2025.
\newblock [Online; accessed 19-February-2025].

\end{thebibliography}

\end{document}